\documentclass[letterpaper, 10 pt, conference]{ieeeconf}  

\IEEEoverridecommandlockouts                              

\usepackage{textcomp}
\usepackage{stfloats}
\usepackage{verbatim}
\usepackage{cite}
\usepackage{xcolor}
\usepackage{mathtools}

\usepackage[colorlinks=true,linkcolor=black,citecolor=black,urlcolor=blue]{hyperref}

\usepackage{microtype}
\usepackage{graphicx}
\usepackage{subfigure}
\usepackage{booktabs} 
\usepackage{bbm}

\usepackage{pgfplots}

\usepackage{amsmath,amssymb,amsfonts}
\usepackage{xurl}
\usepackage{stackengine}
\usepackage{tikz}
\usetikzlibrary{decorations.pathreplacing}
\usetikzlibrary{positioning,arrows.meta,quotes}
\usetikzlibrary{shapes,snakes}
\usetikzlibrary{bayesnet}
\tikzset{>=latex}
\tikzstyle{plate caption} = [caption, node distance=0, inner sep=0pt, below left=5pt and 0pt of #1.south]

\usepackage[normalem]{ulem}
\usepackage{multirow}

\title{\LARGE \bf
Interactive Car-Following: Matters but NOT Always
}

\author{Chengyuan Zhang$^{1}$, Rui Chen$^{2}$, Jiacheng Zhu$^{3}$, Wenshuo Wang$^{1}$, Changliu Liu$^{2}$ and Lijun Sun$^{1\dagger}$
\thanks{$^{1}$Department of Civil Engineering, McGill University, Quebec, Canada.}%
\thanks{$^{2}$Robotics Institute, Carnegie Mellon University, Pennsylvania, USA.}%
\thanks{$^{3}$Department of Mechanical Engineering, Carnegie Mellon University, Pennsylvania, USA.}%
\thanks{*Chengyuan Zhang is currently a visiting student researcher at CMU Robotics Institute. This work is done during his visiting.}%
\thanks{$^{\dagger}$Corresponding author: Lijun Sun ({\tt\small lijun.sun@mcgill.ca})}%
}

\begin{document}

\maketitle
\thispagestyle{empty}
\pagestyle{empty}

\begin{abstract}
Following a leading vehicle is a daily but challenging task because it requires adapting to various traffic conditions and the leading vehicle's behaviors. However, the question \textit{`Does the following vehicle always actively react to the leading vehicle?'} remains open. To seek the answer, we propose a novel metric to quantify the interaction intensity within the car-following pairs. The quantified interaction intensity enables us to recognize interactive and non-interactive car-following scenarios and derive corresponding policies for each scenario. Then, we develop an interaction-aware switching control framework with interactive and non-interactive policies, achieving a human-level car-following performance. The extensive simulations demonstrate that our interaction-aware switching control framework achieves improved control performance and data efficiency compared to the unified control strategies. Moreover, the experimental results reveal that human drivers would not always keep reacting to their leading vehicle but occasionally take safety-critical or intentional actions --- interaction matters but not always.
\end{abstract}

\section{Introduction}
\begin{figure}[t]
    \centering
    \includegraphics[width=\linewidth]{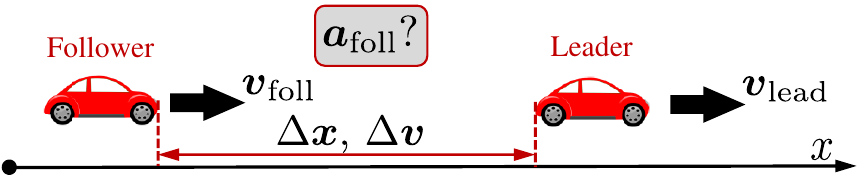}
    \caption{Car-following problem illustration.}
    \label{fig:cf}
\end{figure}

Autonomous driving systems promise to revolutionize our transport networks by enhancing safety, efficiency, and convenience. One challenging task is to follow a leading vehicle (i.e., leader) like a human driver --- a seemingly simple yet intricate operation (as illustrated in Fig.~\ref{fig:cf}). The complexity arises from the need to adapt to ever-changing traffic conditions and the diverse behaviors of the leader. In general, there are two types of car-following models used for autonomous vehicles: one is stimulus-response-based, and the other is learning-based.

Most car-following models are developed with the assumption that the following vehicle (i.e., follower) \textit{always} actively response to the changes of environment states (e.g., leader's speed and position, as the stimulus), such as the Newell's model \cite{newell1961nonlinear}, optimal velocity model \cite{bando1995dynamical}, and intelligent driver model (IDM) \cite{treiber2000congested}. These models already encoded some prior knowledge and thus do not require much data for calibration \cite{zhang2022bayesian, punzo2021calibration}. However, these models heavily rely on the stimulus-response assumption: The follower's reaction is sensitive to the leader's instantaneous action. Our driving experience indicates that in natural traffic settings, the follower's response is scenario-dependent --- drivers will take strong reactions in interactive scenarios but weak reactions in non-interactive scenarios. This is why these stimulus-response-based car-following models might fail to capture varying traffic environments. 

Many learning-based models trained with a large amount of data are developed to capture driving behaviors in diverse driving environments, such as Gaussian mixture models (GMMs) \cite{angkititrakul2009evaluation}, deep neural networks \cite{wang2017capturing}, and deep reinforcement learning \cite{zhu2018human}. With the power of big data, such methods could cover both interactive and non-interactive scenarios using a unified model.
However, training a unified model often demands an intricate architecture and extensive training data to depict the car-following behavior in different traffic scenarios, posing significant challenges in practice.
For instance, sufficiently representing the complex and stochastic driving environment requires tons of real-world data to approximate the true distribution of the behaviors \cite{feng2021intelligent}. Moreover, the low proportion of interactive behaviors in all driving behaviors could lead to a biased model due to the imbalanced data \cite{yan2023learning}.

To overcome the limitations of stimulus-response and learning-based car-following models, we argue that the follower does not always react to the leading vehicle but occasionally takes a safety-critical or intentional response. To this end, we introduce and design a new metric to quantify the interactions, i.e., the intensity of interactions within car-following pairs. Quantified interactions enables us to recognize between interactive and non-interactive scenarios, providing the basis for developing interactive and non-interactive policies. To verify the effectiveness of the interaction intensity metric, we propose an interaction-aware switching control framework, which allows the follower to adaptively switch between the two policies. Extensive simulations indicate that our interaction-aware switching control framework outperforms traditional unified car-following strategies regarding control performance and data efficiency.

In summary, our contributions are as follows:
\begin{enumerate}
    \item We introduce interaction intensity as a quantifiable metric to determine the intensity level of interaction within the car-following pairs (Section \ref{quantify_int}).
    \item We develop an interaction-aware switching control framework by leveraging interaction intensity to decide to switch between interactive and non-interactive policies (Section \ref{swtiching_control}).
    \item We preliminary demonstrate that the follower would not always actively react to the leader but occasionally take safety-critical or intentional actions (Section \ref{exp}).
\end{enumerate}

\section{Interaction Quantification}\label{quantify_int}
Quantifying the intensity of interaction is a critical cornerstone for our interaction-aware switching control, and a quantifiable definition of social interaction in traffic scenarios can be \cite{wang2022social}: \textit{`A dynamic sequence of acts that mutually consider the actions and reactions of individuals through an information exchange process between two or more agents to maximize benefits and minimize costs.'} This definition implies that checking the influences of human drivers on each other can identify the absence and presence of human interactions. Here we assume that the follower's decisions can be formulated as a probability distribution. And we also assume that the leader's action may directly have a real-time impact on the follower's reaction, which is reflected by the shifting in the follower's probability distribution. Therefore, we are interested in estimating the interaction intensity $\mathcal{I}$ between the leader-follower pair by measuring to what extent the probability distribution shifts upon the leader's actions.

\subsection{Interaction Influence Formulation}
We denote $\boldsymbol{a}_{\mathrm{foll}}^{1:t}$ and $\boldsymbol{\hat{a}}_{\mathrm{foll}}^{t+1:t+\Delta T}$ as historical and future (distinguished by the hat symbol) action sequences of the follower, respectively. We then denote the state sequences $\boldsymbol{s} = [\boldsymbol{v}_{\mathrm{foll}}^{1:t}, \Delta\boldsymbol{v}^{1:t}, \Delta\boldsymbol{x}^{1:t}]$ as a concatenation of the follower's speed and relative speed and distance. For the simplicity of notation, we will omit the superscripts in the following. There are mainly two parts of information conveyed by $\boldsymbol{s}$, the leader's motion state $\boldsymbol{s}_{\mathrm{lead}}=[\Delta\boldsymbol{v}, \Delta\boldsymbol{x}]$ and the follower's motion state $\boldsymbol{s}_{\mathrm{foll}}=\boldsymbol{v}_{\mathrm{foll}}$.  The intuition behind designing the interaction metric $\mathcal{I}$ is to investigate the influences of the leader's state $\boldsymbol{s}_{\mathrm{lead}}$ on the follower's future action $\boldsymbol{\hat{a}}_{\mathrm{foll}}$,
\begin{equation}
    \mathcal{I}(\boldsymbol{a}_{\mathrm{foll}},\boldsymbol{s})\coloneqq\mathcal{D}\big(\underbrace{p(\boldsymbol{\hat{a}}_{\mathrm{foll}}|\boldsymbol{s}_{\mathrm{foll}},\boldsymbol{s}_{\mathrm{lead}},\ast)}_{\text{conditional dist.}\ f}||\underbrace{p(\boldsymbol{\hat{a}}_{\mathrm{foll}}|\boldsymbol{s}_{\mathrm{foll}},\ast)}_{\text{marginal dist.} \ g}\big),
\end{equation}
with a conditional behavior model $ p(\boldsymbol{\hat{a}}_{\mathrm{foll}}|\boldsymbol{s}_{\mathrm{foll}},\boldsymbol{s}_{\mathrm{lead}},\ast)$, and a marginalized conditional behavior model $ p(\boldsymbol{\hat{a}}_{\mathrm{foll}}|\boldsymbol{s}_{\mathrm{foll}},\ast)$, where $\ast$ represents the conditions on the action history $\boldsymbol{a}_{\mathrm{foll}}$ and the model parameters. We use a distance-based measure $\mathcal{D}(\cdot||\cdot)$ to evaluate the distance between these two probability distributions, indicating the influences of the leader's states on the follower's future action. Computing the distance depends the probabilistic formulations, and we will parameterize these two models using Gaussian mixture regression (GMR) since it allows for conditionalization and marginalization.

\subsection{Conditional and Marginal Behavior Models}
GMR is widely-used for multivariate nonlinear regression modeling \cite{ghahramani1993supervised} and car-following behavior modeling \cite{wang2019learning, lefevre2015learning}. One fundamental step of GMR is modeling the generative processes of the car-following data as a GMM parameterized by $\boldsymbol{\theta}$, i.e., the joint distribution $p_{\boldsymbol{\theta}}(\boldsymbol{a}_{\mathrm{foll}},\boldsymbol{\hat{a}}_{\mathrm{foll}},\boldsymbol{s}_{\mathrm{foll}},\boldsymbol{s}_{\mathrm{lead}})$ is formulated as GMM, from which we can derive the conditional distribution (which is still a GMM) to approximate the nonlinear function 
\begin{equation}
f: (\boldsymbol{s}_{\text{foll}},\boldsymbol{s}_{\text{lead}},\boldsymbol{a}_{\text{foll}})\mapsto \boldsymbol{\hat{a}}_{\text{foll}}
\end{equation}
in regression tasks. By taking marginalization, one can derive $g = p_{\boldsymbol{\theta}}(\boldsymbol{\hat{a}}_{\mathrm{foll}}|\boldsymbol{s}_{\mathrm{foll}},\boldsymbol{a}_{\mathrm{foll}})$ as another GMM.

\subsection{Quantifying Decision Shifting}
Two popular methods used for measuring the dissimilarity between two probability distributions are the Jenson-Shannon (JS) divergence and Wasserstein distance. The above section indicates that both $f$ and $g$ are mixtures of $K$ Gaussian components as
\begin{subequations}
\begin{align}
f(\boldsymbol{x}) & = \sum_{i=1}^{K} \pi_i^f \mathcal{N}(\boldsymbol{x} | \boldsymbol{\mu}_i^f, \boldsymbol{\Sigma}_i^f) \\
g(\boldsymbol{x}) & = \sum_{j=1}^{K} \pi_j^g \mathcal{N}(\boldsymbol{x} | \boldsymbol{\mu}_j^g, \boldsymbol{\Sigma}_j^g)
\end{align}
\end{subequations}
where ($\pi_i^f$, $\pi_j^g$), ($\boldsymbol{\mu}_i^f$, $\boldsymbol{\mu}_j^g$), and ($\boldsymbol{\Sigma}_i^f$, $\boldsymbol{\Sigma}_j^g$) are the weights, means, and covariance, respectively. To simply notations, we denote the Gaussian distributions $\mathcal{N}(\boldsymbol{x} | \boldsymbol{\mu}_i^f, \boldsymbol{\Sigma}_i^f)$ as $\mathcal{N}_i^f$ and $\mathcal{N}(\boldsymbol{x} | \boldsymbol{\mu}_j^g, \boldsymbol{\Sigma}_j^g)$ as $\mathcal{N}_j^g$. In what follows, we will introduce the JS divergence and Wasserstein distance of $f$ and $g$ for quantifying the interactions. 

\subsubsection{JS divergence} The JS divergence between two probability distributions $f$ and $g$ is defined as
\begin{equation}
    \mathcal{D}_{\text{JS}}(f, g) = \frac{1}{2} \mathcal{D}_{\text{KL}}(f || h) + \frac{1}{2} \mathcal{D}_{\text{KL}}(g || h),
\end{equation}
where $\mathcal{D}_{\text{KL}}(f || g)$ is the Kullback-Leibler (KL) divergence, and $h(\boldsymbol{x}) = \frac{1}{2}(f(\boldsymbol{x}) + g(\boldsymbol{x}))$. Note that the KL divergence between two GMMs is generally intractable, and we use Monte Carlo sampling to approximately estimate KL \cite{hershey2007approximating}.

\subsubsection{Wasserstein distance}The $p$-th order Wasserstein distance between two GMMs is expressed as
\begin{equation}
    W_p(f, g) = \left( \min_{\gamma \in \Gamma(f, g)} \sum_{i=1}^{K} \sum_{j=1}^{K} \gamma_{ij} d_p(\mathcal{N}_i^f, \mathcal{N}_j^g) \right)^{1/p},
\end{equation}
where $\Gamma(f, g)$ is the set of all couplings between the two distributions, $\gamma_{ij}$ is the elements of the optimal coupling matrix, and $d_p(\mathcal{N}_i^f, \mathcal{N}_j^g)$ is the $p$-th order distance between the Gaussian components $\mathcal{N}_i^f$ and $\mathcal{N}_j^g$. The $p$-th order Wasserstein distance for Gaussians can be calculated by
\begin{equation}
    d_p(\mathcal{N}_i^f, \mathcal{N}_j^g) = \Vert \mu_i^f - \mu_j^g \Vert^p + \mathcal{B}_p(\Sigma_i^f, \Sigma_j^g),
\end{equation}
where $\mathcal{B}_p(\Sigma_i^f, \Sigma_j^g)$ is a Bures-like distance between the covariance matrices. The Bures-like distance is generally not available in closed-form, but there exist approximations and optimization techniques to compute it as well \cite{chen2018optimal, delon2020wasserstein}.

\section{Interaction-Aware Switching Control-Based Car-Following Model}
\label{swtiching_control}
Instead of directly training a unified car-following policy using all car-following behavior data, we propose to utilize several interaction-aware sub-policies, and switch between them based on interaction intensity, i.e., the value of $\mathcal{I}$. To this end, we need to tackle two challenges: (a) construction of the sub-policies and (b) design of the switching mechanism.
To acquire sub-policies, we classify car-following behaviors into \textit{interactive} and \textit{non-interactive} ones according to the value of interaction intensity $\mathcal{I}$, and train a separate model for each. This allows us to exploits the benefits of training \textit{ad hoc} models and only use minimal data to overcome the data imbalance problem. This aligns with the fact that intense interactive behaviors are rare in naturalistic data. The proposed switching mechanism is the core of this framework, and we will explicitly elaborate it in the following.

Switching control consists of different control policies for various operating modes (e.g., interactive or non-interactive) the car-following behaviors. Depending on the current mode of the leader-follower pair, an appropriate control policy is selected and applied. For instance, given an interactive car-following policy $\pi_{\text{int}}$ and a non-interactive policy $\pi_{\text{non}}$, a high-level supervisory logic $\psi$ is used to decide which one to apply at each moment.

For the follower in a car-following pair, given the current state, we seek to select the control policy according to the interaction intensity $\mathcal{I}$, with the switch logic function
\begin{equation}\label{hard_eqn}
    \psi(\mathcal{I}) = \begin{cases}\text{select}\,\pi_{\text{int}},\, \text{if } \, \mathcal{I}>\mathcal{I}_0,\\
    \text{select}\,\pi_{\text{non}},\, \text{if } \, \mathcal{I}\leq\mathcal{I}_0,
    \end{cases}
\end{equation}
where $\mathcal{I}_0$ is a intensity threshold, above which is considered as intense interaction intensity.

However, one should note that switching between controllers can cause transient effects or stability issues, especially if the controllers are not designed with smooth transitions in mind. Therefore, we developed a soft switching scheme to model the the supervisory logic $\psi$ as a mixing of both two policies

\begin{equation}\label{soft_eqn}
    \pi_{\text{switch}} =  \psi(\mathcal{I}) \pi_{\text{int}} + (1-\psi(\mathcal{I})) \pi_{\text{non}},
\end{equation}
where $\psi(\mathcal{I}) = \sigma\left(\frac{\mathcal{I}-\mathcal{I}_0}{\beta}\right)$, in which $\sigma$ represents the sigmoid function and $\beta$ is a scaling factor. The intuition behind this setting is putting more weights on the interaction policy $\pi_{\text{int}}$ when encounters an intensely interactive situation, while maintaining a smooth transition between the two policies.


\section{Experiment Results and Analysis}\label{exp}
\subsection{Dataset and Experiment Settings}
We use the HighD dataset \cite{krajewski2018highd}, a high-resolution trajectory data collected using drones. It has $60$ video recordings, logged with the sampling frequency of $25$ Hz on several German highway sections with a length of $420$ m. To simplify our data, we downsample the original dataset to a smaller set with sampling frequency of $5$ Hz (i.e., the time step between consecutive data points is $0.2$ sec.) In each recording, the trajectories, velocities, and accelerations are measured and estimated. We follow the same data processing procedures as in \cite{zhang2021spatiotemporal} to transform the data into a new coordinate system. We extract informative car-following pairs according to \cite{zhang2022bayesian}.

\begin{figure}[t]
    \centering
    \subfigure{
        \centering\includegraphics[width=.465\linewidth]{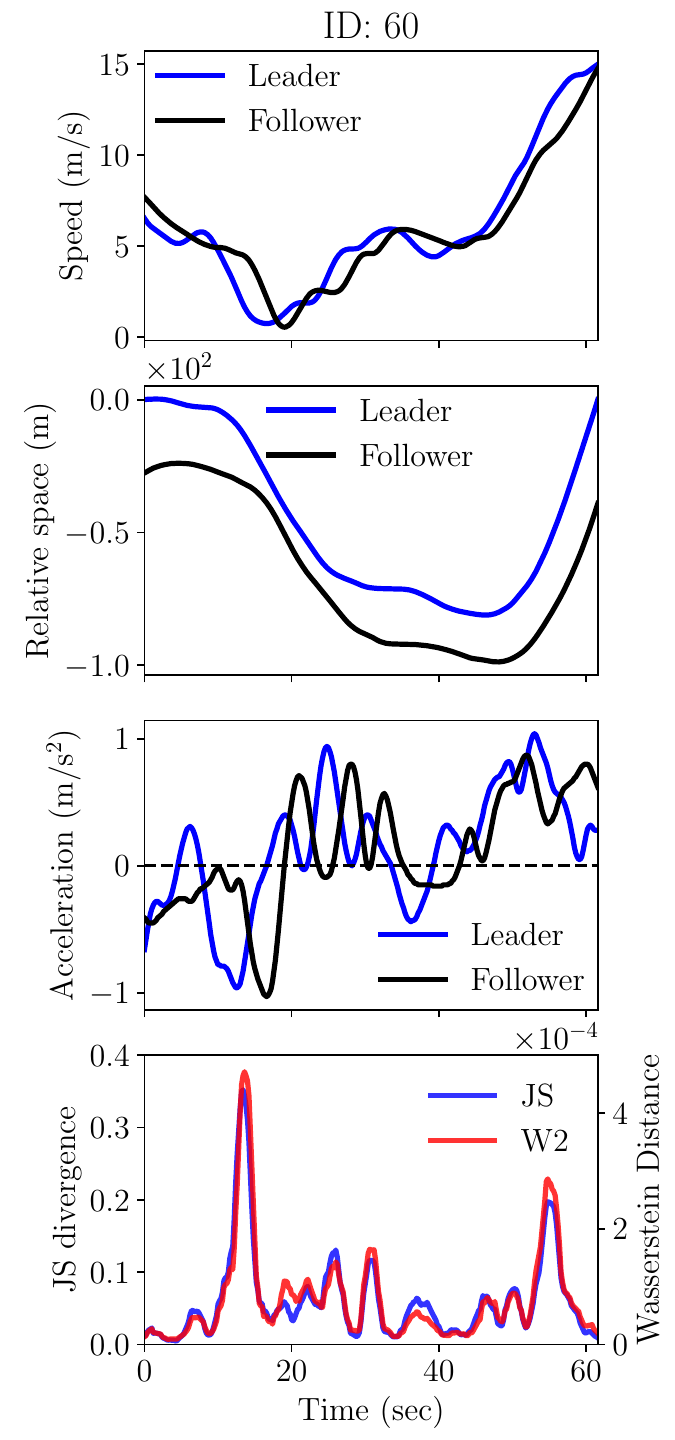}
    }
    \subfigure{
        \centering\includegraphics[width=.465\linewidth]{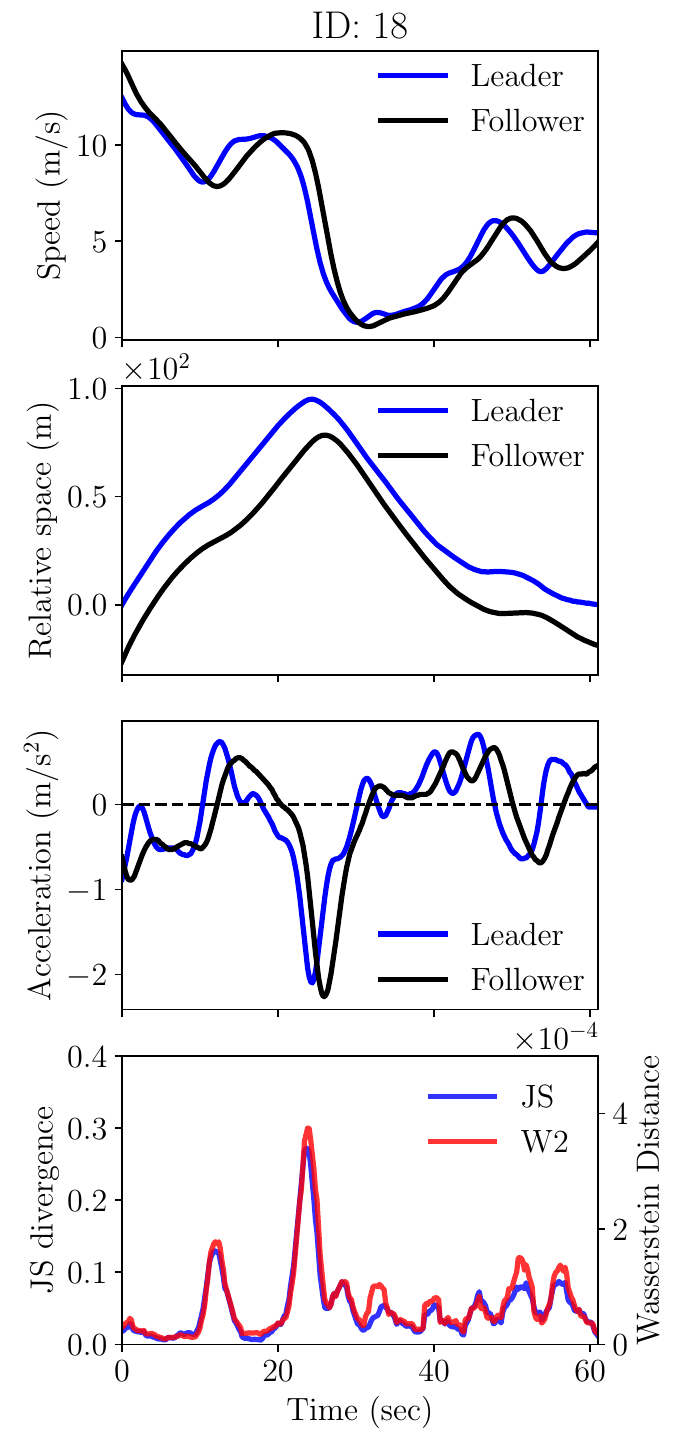}
    }
    \caption{The motion profiles and the quantified interaction intensity of two car-following pairs. Note that the second row shows the trajectories at the point view of another constant-mean-speed `observing' vehicle. Therefore, the relative space of the leader at the beginning and the ending are both zeros.}
    \label{fig:quantified_results_two}
\end{figure}

\subsection{Learning Car-Following Policies}
\subsubsection{Quantification Results}
Here we set the historical time horizon $t=1$ sec and the future prediction time horizon $\Delta T=0.6$ sec, and train the GMR on $200$ randomly selected car-following pairs. Then we randomly selected another $20$ pairs to test the control policy. Basically, given the observations of $(\boldsymbol{s}_{\text{foll}},\boldsymbol{s}_{\text{lead}},\boldsymbol{a}_{\text{foll}})$, we evaluate $\mathcal{D}_{\text{JS}}(f||g)$ and $W_2(f,g)$ at any time step. The quantified interactions of two random car-following pairs are shown in Fig.~\ref{fig:quantified_results_two}. The bottom indicates that the interactions quantified by JS divergence and 2-Wasserstein (W2) distances have almost the same trends, except that their values have different scales. Therefore, in the following parts of this paper, we will not distinguish between the different quantification methods but only use JS divergence in default.

\begin{figure}[t]
    \centering
    \includegraphics[width=.95\linewidth]{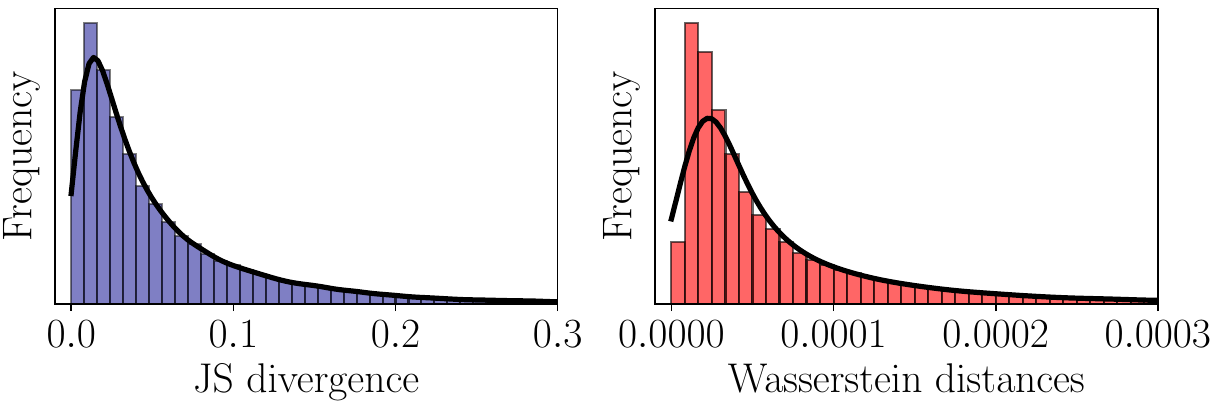}
    \caption{Histograms of the quantified interaction intensity.}
    \label{fig:histogram}
\end{figure}

For an individual's driving behavior, the intense interaction indicates that the follower takes a strong reaction to the leader's action. For instance, the leader takes abrupt braking or the leader stops pushing the gas pedal after a rapid acceleration. To better understand the car-following interaction from the population level, we visualize the histogram of the quantified interaction intensity on all of the available car-following pairs in Fig.~\ref{fig:histogram}. Notice that the human drivers tend to drive without intense interaction in most of the time.

\subsubsection{Interactive/Non-interactive Data Sampling}
\begin{figure*}[t]
    \centering
    \subfigure{
        \centering\includegraphics[width=.31\linewidth]{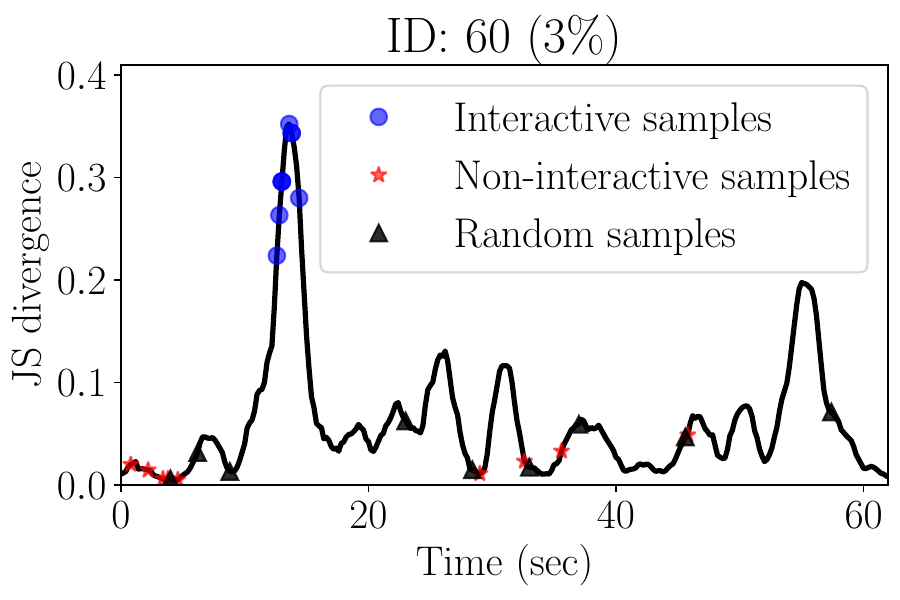}
    }
    \subfigure{
        \centering\includegraphics[width=.31\linewidth]{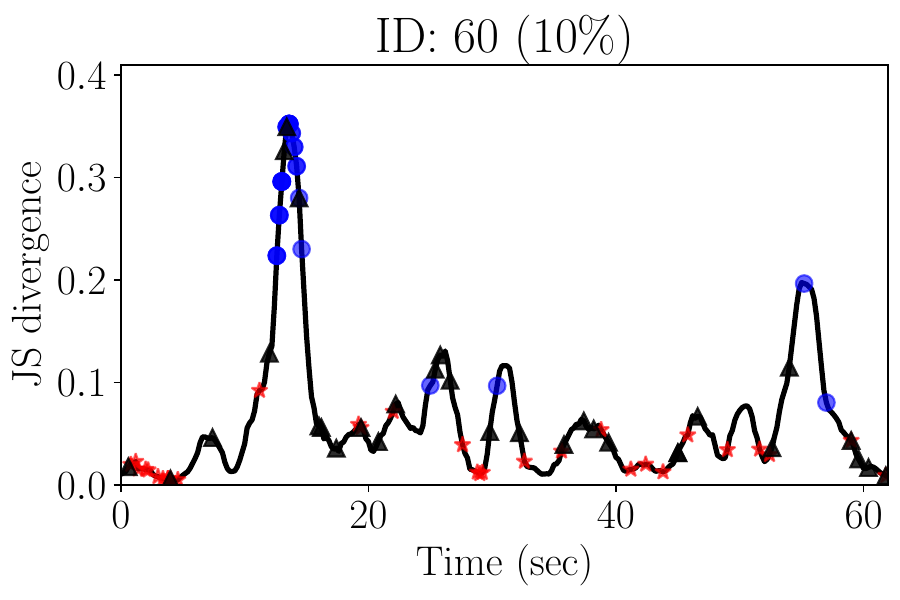}
    }
    \subfigure{
        \centering\includegraphics[width=.31\linewidth]{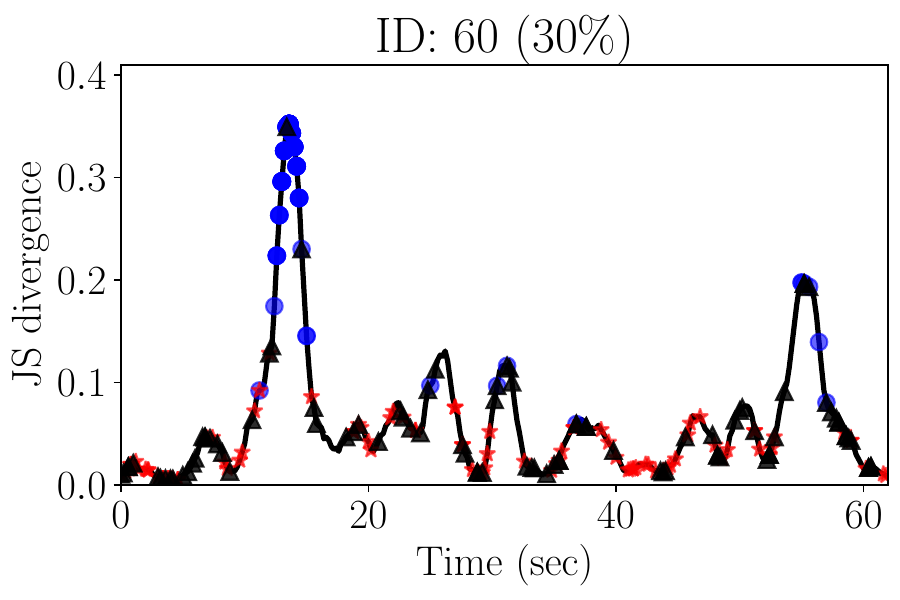}
    }
    \caption{The interactive and non-interactive samples. The blue dots represent interactive samples, the red stars denote non-interactive samples, and the black triangles are random samples. The percentage in parentheses represents the amount of samples.}
    \label{fig:data_importance_samples}
\end{figure*}

Recall that we evaluate the interaction intensity $\mathcal{I}$ at any time step in Fig.~\ref{fig:quantified_results_two}, it is straightforward to sample interactive/non-interactive data based on the interaction intensity. We illustrate this intuition with Fig.~\ref{fig:data_importance_samples}, where $3\%$, $10\%$, and $30\%$ data are sampled from the original trajectory.

\subsubsection{Learning Interactive/Non-interactive Models}
To verify the performance under data insufficient cases, we only use $3\%$ interactive/non-interactive samples from a full trajectory (see Fig.~\ref{fig:data_importance_samples}) to obtain $\pi_{\text{int}}$ and $\pi_{\text{non}}$, respectively. Here we use the IDM \cite{treiber2000congested} as the car-following policy, and adopt the Bayesian calibration method proposed in \cite{zhang2022bayesian} to identify the IDM parameters' distribution, from which we could draw many sets of IDM parameters.  In addition, another IDM $\pi_{\text{rand}}$ is calibrated as the baseline with $6\%$ randomly sampled data, which contains both interactive and non-interactive samples at random.

\subsection{Simulations with Interaction-Aware Switching Control}
\begin{figure*}[t]
    \centering
    \subfigure[Hard-switching control (ID: 03).]{
        \centering\includegraphics[width=.48\linewidth]{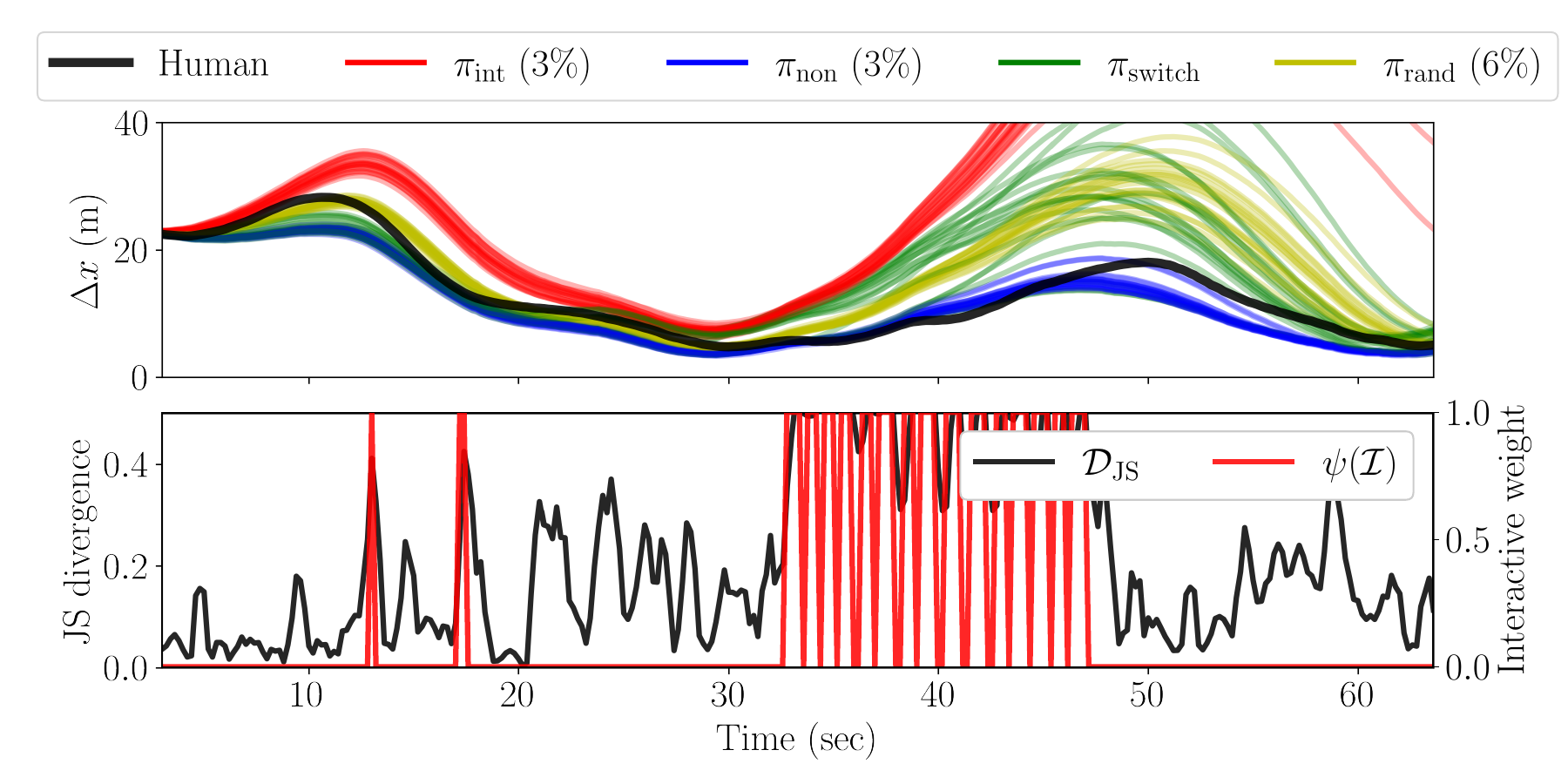}
    }
    \subfigure[Hard-switching control (ID: 232).]{
        \centering\includegraphics[width=.48\linewidth]{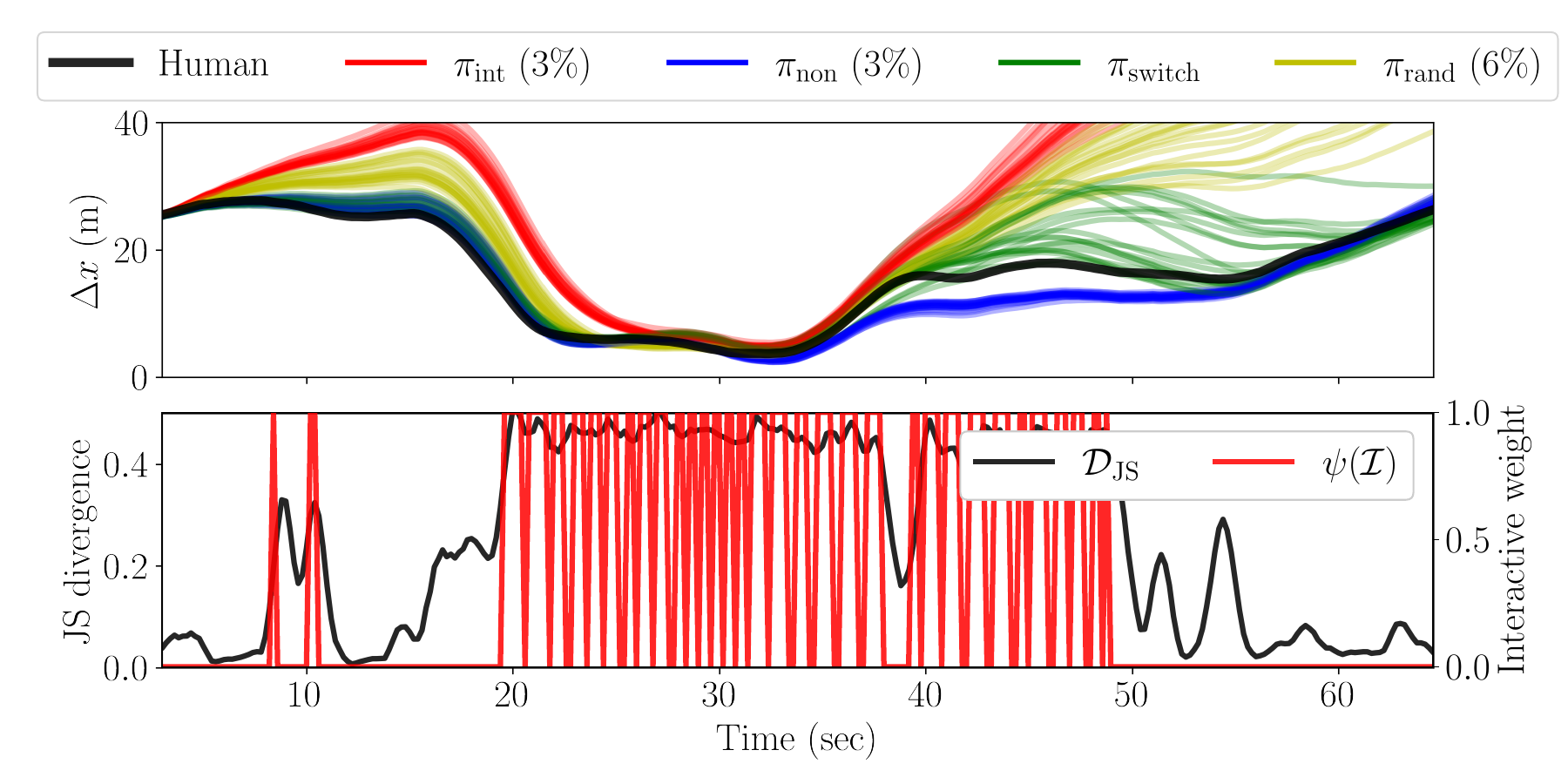}
    }\\
    \subfigure[Soft-switching control (ID: 03).]{
        \centering\includegraphics[width=.48\linewidth]{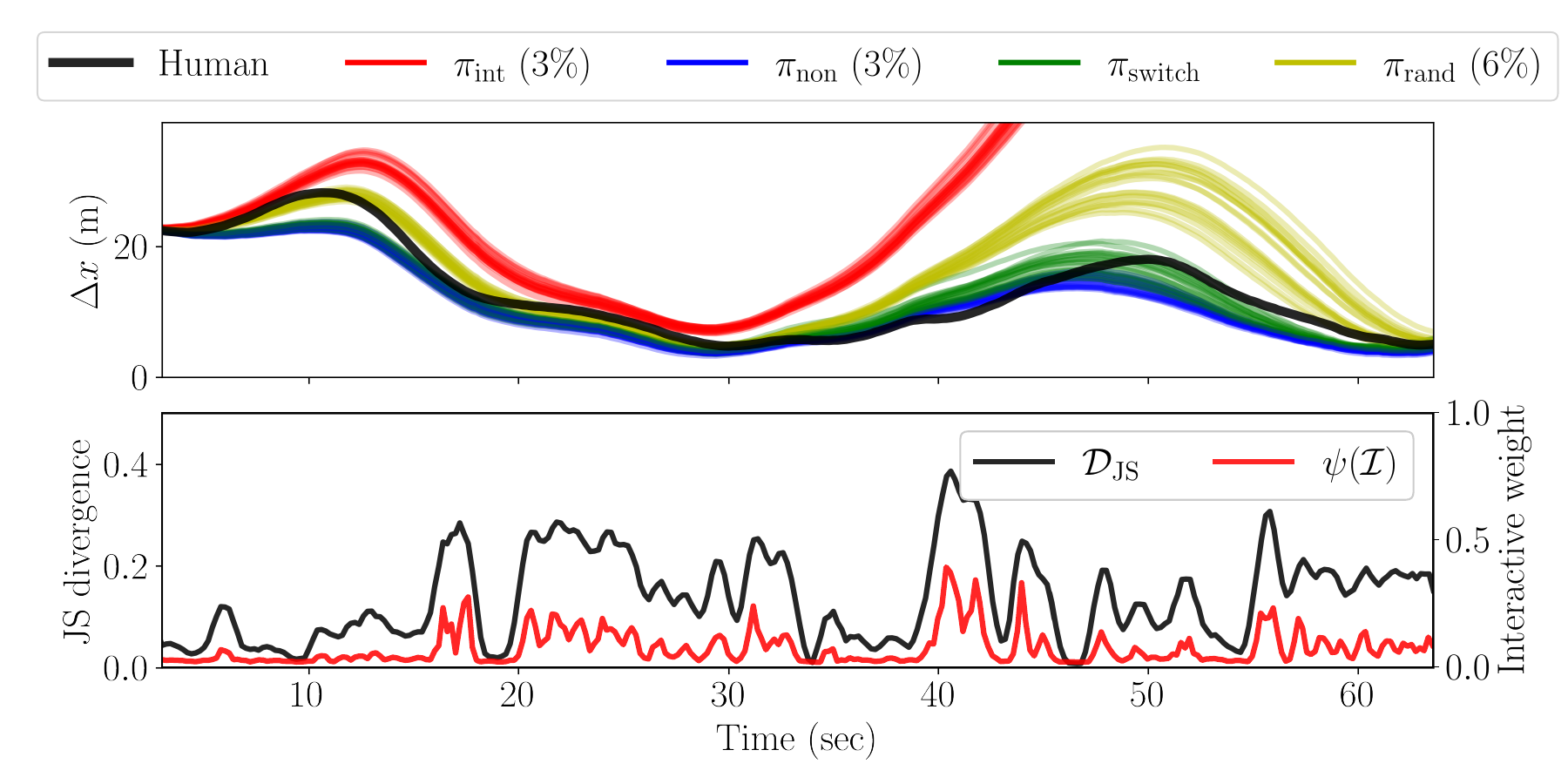}
    }
    \subfigure[Soft-switching control (ID: 232).]{
        \centering\includegraphics[width=.48\linewidth]{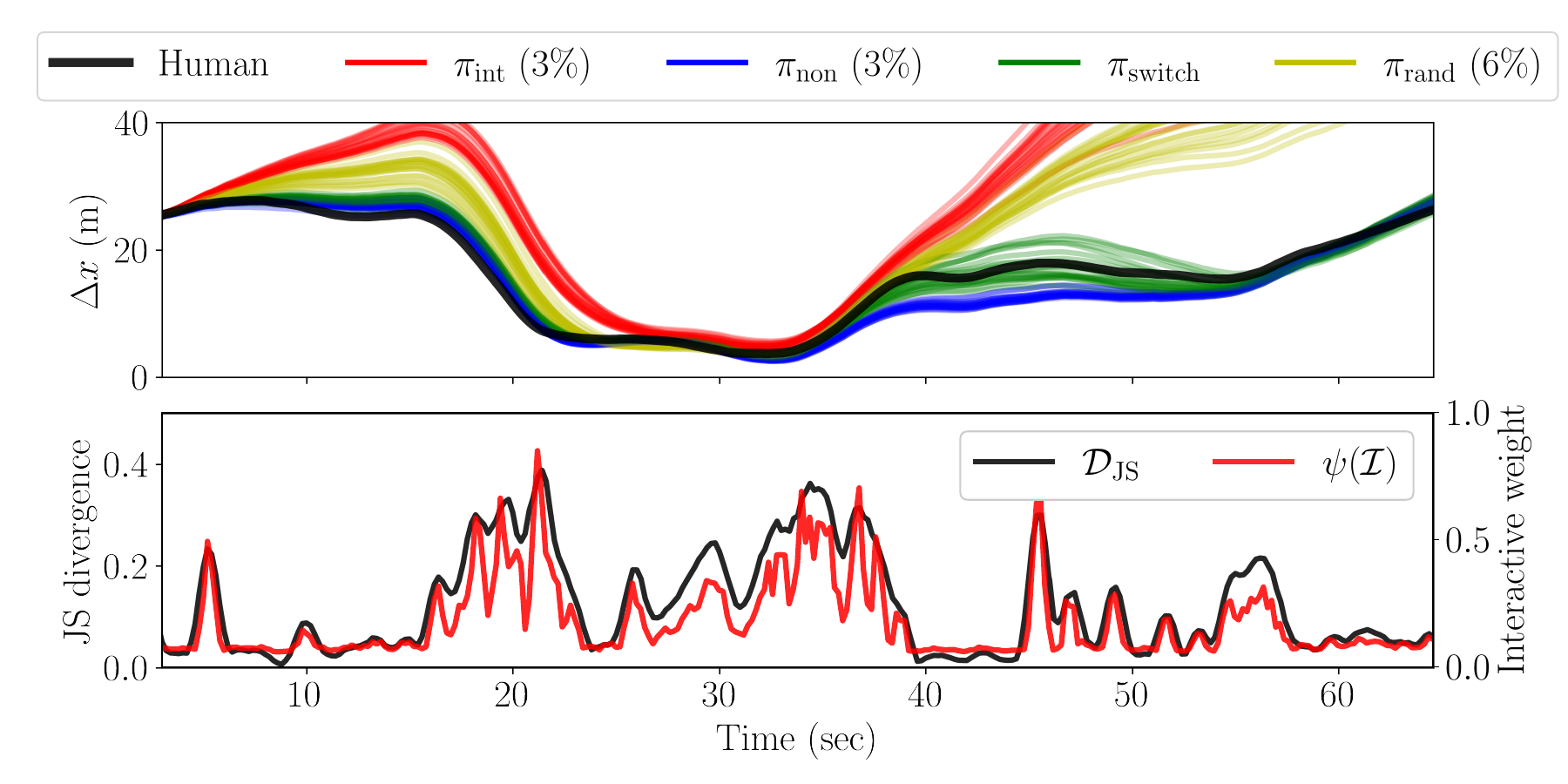}
    }
    \caption{The simulated trajectories and quantified interaction intensity of two followers controlled by hard/soft-switching policies, respectively. The upper rows illustrate the simulated trajectories. The lower rows represent the quantified interaction intensity (left axis) and the interactive weights $\psi(\mathcal{I})$ (right axis), which correspond to one of the green lines.}
    \label{fig:sim}
\end{figure*}

In this part, we evaluate and compare the performances of different control polices in simulation. Specifically, the follower takes actions by a specific control policy to follow a human leader. We run the simulation with the same initial states for several times, and the comparison of $\pi_{\text{int}}$, $\pi_{\text{non}}$, $\pi_{\text{rand}}$, and the hard-switching policy $\pi_{\text{switch}}$ for two car-following pairs are illustrated in Fig.~\ref{fig:sim}(a) and Fig.~\ref{fig:sim}(b). The interactive policy $\pi_{\text{int}}$ learns to take safety-critical actions in scenarios with intense interactions, such as collision avoidance; while the non-interactive policy $\pi_{\text{non}}$ learns to follow the leader and reach the target speed. Therefore, the results indicate that $\pi_{\text{int}}$ behaves too conservative; it tends to keep a low speed with a large space headway. $\pi_{\text{non}}$ usually keeps a short space headway and actively follows the leader; and $\pi_{\text{rand}}$ seems to be a compromise between the two strategies. In general, $\pi_{\text{switch}}$ takes the characteristics of both $\pi_{\text{int}}$ and $\pi_{\text{non}}$ by switching between the `actively following' mode and the `avoiding collision' mode according to the interaction intensity $\mathcal{I}$. As a comparison, the results of $\pi_{\text{rand}}$ indicate that a parsimonious model cannot be well-calibrated with so limited data.

As mentioned previously, switching between controllers can cause transient effects or stability issues, especially if the controllers are not designed with smooth transitions, see the jumping interactive weights at the bottom parts in Fig.~\ref{fig:sim}(a) and Fig.~\ref{fig:sim}(b). Since the hard-switching mechanism is set as a step function in (\ref{hard_eqn}), the stability of the system is very sensitive to the switching points. It requires carefully tuning of the intensity threshold $\mathcal{I}_0$ in application. Therefore, with a fixed $\mathcal{I}_0$, although we can find some results under hard-switching control that replicate the human-driver trajectories pretty well, the specific threshold apparently cannot fit all of the sets of IDM parameters drawn from the learned policies. Therefore, a soft-switching control policy is significant.

Setting a sigmoid switching function instead of the step function could be an effective solution to this issue. To illustrate, we evaluate the soft-switching control policy based on (\ref{soft_eqn}). The results are demonstrated in Fig.~\ref{fig:sim}(c) and Fig.~\ref{fig:sim}(d). Here we quantitatively evaluate the simulated trajectories for 7 distinct car-following pairs in Table~\ref{tab:errors}. The performance metric used is the root-mean-square error (RMSE) of the spatial headway ($\Delta x$) and safety measure. For the safety measure, we evaluate how far the simulated trajectories are closer to the leader than the human driver's trajectories, thus, a lower RMSE indicates a safer policy. Each cell in the table contains the mean and standard deviation of the RMSE over multiple simulation runs. The bold numbers indicate the lowest RMSE value for each car-following pair, which represents the best-performing policy for that scenario. Given that $\pi_{\text{int}}$ is too conservative and it keeps a large spacing behind the leader, thus we didn't evaluate its safety measure in Table~\ref{tab:errors}. Overall, the table demonstrates the superiority of the soft-switching policy $\pi_{\text{switch}}$ in most scenarios, achieving a good balance between actively following the leader and ensuring safety, which is the ultimate goal of our proposed approach.

\begin{table*}[!h]
    \footnotesize
    \centering
    \caption{Evaluations of different control policies on $7$ car-following pairs.}
    \begin{tabular}{c|c|c|c|c|c|c|c}
    \toprule
        $\mathrm{RMSE}(\Delta x)$ & $\#03$ & $\#232$ & $\#14$ & $\#23$ & $\#153$ & $\#144$ & $\#81$\\
    \midrule
        $\pi_{\text{int}}$ (3\%) & $29.35\pm2.68$   & $25.37\pm4.48$ & $16.26\pm0.63$ & $8.23\pm0.41$ & $13.28\pm1.77$ & $7.89\pm0.68$ & $6.12\pm0.35$\\
        $\pi_{\text{non}}$ (3\%) & $2.66\pm0.22$ & $2.36\pm0.19$ & $\mathbf{1.96\pm0.19}$ & $\mathbf{1.17\pm0.04}$ & $1.47\pm0.12$ & $1.70\pm0.05$ & $\mathbf{1.37\pm0.37}$\\
        $\pi_{\text{switch}}$  & $\mathbf{2.43\pm0.24}$  & $\mathbf{1.26\pm0.21}$ & $2.33\pm0.59$ & $1.36\pm0.34$ & $\mathbf{1.30\pm0.13}$ & $\mathbf{1.39\pm 0.09}$ & $1.87\pm0.42$\\
        $\pi_{\text{rand}}$ (6\%) & $5.94\pm1.38$ & $14.83\pm2.84$ & $2.04\pm0.38$ & $3.31\pm0.66$ & $1.67\pm0.13$ & $1.44\pm0.07$ & $2.95\pm0.27$\\
    \midrule
        $\mathrm{RMSE(safe)}$ & $\#03$ & $\#232$ & $\#14$ & $\#23$ & $\#153$ & $\#144$ & $\#81$\\
    \midrule
        $\pi_{\text{int}}$ (3\%) & - & - & - & - & - & - & - \\
        $\pi_{\text{non}}$ (3\%) & $2.83\pm0.25$  & $2.65\pm0.28$ & $2.35\pm 0.42$ & $1.29\pm0.05$ & $1.90\pm0.14$ & $2.06\pm0.07$ & $0.52\pm0.33$\\
        $\pi_{\text{switch}}$  & $2.39\pm0.24$ & $1.01\pm0.26$ & $\mathbf{1.12\pm0.45}$ & $\mathbf{0.31\pm0.08}$ & $\mathbf{1.50\pm0.36}$ & $1.94\pm0.19$ & $\mathbf{0.34\pm0.32}$\\
        $\pi_{\text{rand}}$ (6\%) & $\mathbf{0.72\pm0.15}$ & $\mathbf{0.51\pm2.82}$ & $2.31\pm0.53$ & $1.07\pm0.08$ & $1.74\pm0.19$ & $\mathbf{1.64\pm0.17}$ & $0.75\pm0.20$\\
    \bottomrule
    \end{tabular}
    \label{tab:errors}
\end{table*}

\subsection{Discussions and Limitations}
Our interaction-aware switching control method has demonstrated promising results with a loose data requirement in improving the efficiency and performance of car-following control in autonomous vehicles. By quantifying the level of interaction required between vehicles, our approach enables a more adaptive and context-aware control strategy that can handle a wide range of driving scenarios.

In addition, the quantification results in Fig.~\ref{fig:histogram} revealed from the population level that intense interactions are rare events in the car-following task. The results in Fig.~\ref{fig:sim} further confirmed this point that the interactive policy $\pi_{\text{int}}$ is only actively adopted for a small proportion across the whole time horizon. In general, the results validate our hypothesis that not all car-following scenarios require the follower to take interactive reactions with respect to the leader, but safety-critical or intentional actions are occasionally needed; interactive car-following policy matters but not always. This interesting finding is consistent with the intuition that human typically do complex tasks using simple actions \cite{zhang2021spatiotemporal, wang2018driving}. Our results shed some potential insights that social interactions behind overwhelmingly complex human driving behaviors are not always complicated but governed by some simple rules.

However, despite its promising results, several limitations are critical to our approach. First, the proposed quantification method heavily relies on the performance of the car-following behavior model (i.e., GMR). The behavior model is crucial for the success of our approach, as it determines when to switch between the control policies. Our current method for quantifying interaction intensity may not be optimal or universally applicable, and it might require further refinement or adaptation to different driving environments and vehicle dynamics. Second, although our method has shown to reduce transient effects when switching between control policies, ensuring smooth transitions remains a challenge. The design of the interactive and non-interactive policies must take into account the possibility of abrupt changes in control inputs to prevent undesirable effects on the vehicle's stability and passenger comfort. Third, although our approach has shown promising results in the car-following scenarios, its generalization to other urban traffic conditions, vehicle types, and sensor configurations remains to be validated. Additional experiments and evaluations in diverse urban scenarios are worth trying to verify the robustness and reliability of our method.

\section{Conclusions}\label{conclusion}
In this paper, we present a novel interaction-aware switching control method for car-following scenarios in autonomous driving systems. By introducing the concept of interaction intensity as a quantifiable metric, we develop an adaptive control strategy that switches between interactive and non-interactive policies based on the current driving situation. Through extensive simulations, we demonstrate the effectiveness of our interaction-aware switching control method in adapting to different driving scenarios and achieving superior performance compared to unified control strategies. Our results indicate that considering the varying interaction intensities in car-following scenarios can lead to more robust and efficient autonomous vehicle control. Furthermore, the experiments confirmed that human drivers would not always keep reacting to their leading vehicle but occasionally take safety-critical or intentional actions.

Despite its promising results, our approach is preliminary in the choice of interaction intensity metric, transition smoothness between policies, and generalization to other traffic conditions and vehicle types. Future research should focus on extensions over those directions and further refining our method to enhance its robustness and applicability in complex urban traffics. On a broader scale, our framework also provides insights into designing efficient controllers in other robotics tasks, such as human-robot interactions (HRI), especially when a large amount of human data are expensive to collect.





\section*{ACKNOWLEDGMENT}
C. Zhang would like to thank the McGill Engineering Doctoral Awards (MEDA), the Mitacs Globalink Research Award, Fonds de recherche du Québec -- Nature et technologies (FRQNT), and the Natural Sciences and Engineering Research Council (NSERC) of Canada for providing scholarships and funding to support this study.

{\normalem
\bibliographystyle{IEEEtran}
\bibliography{main_v1}}

\end{document}